%% file: inalu.tex
\documentclass{article}
\usepackage{graphicx}
\usepackage[hidelinks]{hyperref}
\usepackage[]{graphicx}
\usepackage{amsmath}
\usepackage[utf8]{inputenc}
\usepackage{algorithmic}
\usepackage{array}
\usepackage{fixltx2e}
\usepackage{dblfloatfix}
\usepackage{url}
\usepackage{filecontents}
\usepackage{epstopdf}
\usepackage{flushend}
\usepackage{xspace}
\usepackage{amssymb}
\usepackage{siunitx}
\usepackage{csvsimple}
\usepackage{filecontents}
\usepackage{pgfplotstable}
\usepackage{booktabs, colortbl}
\usepackage{multirow}
\pgfplotsset{compat=1.12}
\usepackage{amsmath}
\usepackage[normalem]{ulem}
\usepackage{todonotes}
\usepackage[caption=false]{subfig}

\usepackage{siunitx}
\DeclareMathOperator{\sign}{sign}

\csvset{
  autotabularcenter/.style={
    file=#1,
    after head=\csv@pretable\begin{tabular}{*{\csv@columncount}{c}}\csv@tablehead,
    table head=\toprule\csvlinetotablerow\\\midrule,
    late after line=\\,
    table foot=\\\bottomrule,
    late after last line=\csv@tablefoot\end{tabular}\csv@posttable,
    command=\csvlinetotablerow}}

\newcount\pgfplotstableuniqueentry
\def\pgfplotstableadduniquecol#1#2{%
    \pgfplotstablecreatecol[
        create col/assign/.code={%
            \getthisrow{#1}\entry\getnextrow{#1}\nextentry% This and next for comparison
            \ifx\entry\nextentry\relax% Compare
                \xdef\tempentry{\entry}% Save it to some macro
                \pgfkeyssetvalue{/pgfplots/table/create col/next content}{}% Empty cell
                \global\advance\pgfplotstableuniqueentry1\relax%Increment the dup count
            \else%
                \ifnum\the\pgfplotstableuniqueentry>0\relax% If there is a duplicate
                    \advance\pgfplotstableuniqueentry1% increment for multirow
                    \edef\temp{\noexpand%Selectively expand some macros
                    \pgfkeyssetvalue{/pgfplots/table/create col/next content}{%
                        \noexpand%
                        \multirow{-\the\pgfplotstableuniqueentry}{*}{\tempentry}
                    }}\temp% Execute the expanded version
                    \global\pgfplotstableuniqueentry=0%Reset dup counter
                \else%
                    \pgfkeyslet{/pgfplots/table/create col/next content}{\entry}%Do nothing
                \fi%
            \fi%
        }
    ]{u-#1}{#2}%
}

\begin{document}

\date{}

\title{iNALU: Improved Neural Arithmetic Logic Unit}

\author{Daniel Schlör$^1$ \and
Markus Ring$^2$ \and
Andreas Hotho$^1$ \\[1.5em]
$^1$University of Wuerzburg, Germany \\
$^2$University of Applied Sciences and Art Coburg, Germany\\[1.5em]
\{schloer,hotho\}@informatik.uni-wuerzburg.de \\
markus.ring@hs-coburg.de
}
\maketitle             

\begin{abstract}
Neural networks have to capture mathematical relationships in order to learn various tasks. They approximate these relations implicitly and therefore often do not generalize well. The recently proposed Neural Arithmetic Logic Unit (NALU) is a novel neural architecture which is able to explicitly represent the mathematical relationships by the units of the network to learn operations such as summation, subtraction or multiplication.
Although NALUs have been shown to perform well on various downstream tasks, an in-depth analysis reveals practical shortcomings by design, such as the inability to multiply or divide negative input values or training stability issues for deeper networks.
We address these issues and propose an improved model architecture.
We evaluate our model empirically in various settings from learning basic arithmetic operations to more complex functions. Our experiments indicate that our model solves stability issues and outperforms the original NALU model in means of arithmetic precision and convergence.

\end{abstract}
\input{chapters/introduction}

\input{chapters/related}
\input{chapters/nalu}
\input{chapters/experiments}

\input{chapters/conclusion}

\section*{Acknowledgements}
This work was partly funded by the Federal Ministry of Education and Research of
Germany as part of the DeepScan project (01IS18045A) and the Bavarian Ministry of Economic Affairs Regional Development and Energy through the OBELISK project. M. R. was supported by the BayWISS Consortium Digitization. 

\bibliographystyle{abbrv}
\bibliography{inalu}{}
\end{document}

%% file: chapters/introduction.tex
\section{Introduction}
\label{sec:introduction}

Neural networks have achieved great success in various data mining application areas. 
Thereby, different network structures are suitable for different tasks. 
For instance, convolutional neural networks are well suited for image processing while recurrent neural networks are well suited for handling sequential data.  
However, neural networks also face challenges like processing categorical values or calculating specific mathematical operations.  

The presence of mathematical relationships between features is a well-known fact in many financial tasks~\cite{bolton2002statistical,lopez2016paysim}. 
Other examples can be found in the intrusion detection domain. 
For example, some intrusion detection methods count the number of certain events~\cite{garcia2014empirical} or consider some restrictions such as that network packets have a minimum and maximum number of transmitted bytes~\cite{ring2018gan}. A model which is able to capture these relationships explicitly in an automated way is therefore very desirable and can be incorporated in various machine learning tasks. 

\begin{figure}[h!]
	\includegraphics[width=0.6\textwidth]{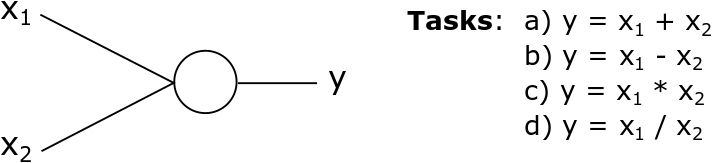}
	\centering
	\caption{Standard mathematical tasks.
	}
	\label{fig:introMathExample}
\end{figure}

\textbf{Problem.}
While neural networks are well suited for many data mining tasks, single neurons often have problems with the calculation of basic mathematical operations~\cite{trask2018neural}. 
This fact can be explained by inspecting the structure of neurons in detail.
The output of a neuron $i$ is the weighted sum of all input signals, an optional bias $b$ and an activation function:

\begin{equation}
output_i=act\left(\big(\sum_{j=1}^{n}x_j \cdot w_j\big) + b_i\right)
\label{frm:neuron}
\end{equation}
The neuron $i$ in Equation~\ref{frm:neuron} receives $n$ input signals $x_j$ which are multiplied by the weights $w_j$.
The parameter $b_i$ represents an optional bias and $act(\cdot)$ is an arbitrary activation function like the identity for a linear or sigmoid for a non-linear neuron. 
This allows neurons to assign different weights to different input features. 
Further, linear neurons are able to add (or subtract) different inputs by setting their corresponding weights to $1$ (or $-1$), see tasks a) and b) in Figure~\ref{fig:introMathExample}.
However, activation functions, weights and bias allow neurons only to approximate the result of multiplications and divisions in their training range, since the output is the weighted sum of all inputs.
Consequently, they can't solve multiplication and division tasks for values outside the training range (see tasks c) and d) in Figure~\ref{fig:introMathExample}).

Trask et al.~\cite{trask2018neural} show empirically that artificial neurons have especially difficulties with extrapolation of mathematical operations and present the \textit{Neural Arithmetic Logic Units} (NALU) to address this problem.
However, the NALU is only able to calculate non-negative results for multiplication and division by design. Madsen and Johansen~\cite{madsen2019measuring} further show that the NALU is not able to learn division reliably and often fails to converge to the desired weights.

\textbf{Objective.}
Inspired by the NALU, we want to improve the architecture to address the above mentioned problems. 
Our focus lies on processing negative values and improving extrapolation by forcing the internal weights to intended values.

\textbf{Contribution.}
In this paper, we propose iNALU as improvement of the NALU  architecture \cite{trask2018neural}.
First of all, we add another path to allow multiplication and division with mixed-signed inputs. 
Further, we propose an input independent implementation of the gate, switching between the summative and multiplicative path. 
Based on empirical observations, we add regularization to the training procedure to prevent approximation of the results due to unwanted combination of mathematical operations. 
Then, a maximum function for the multiplicative path is introduced to avoid too large values (infinity) for deep networks with several hidden layers and many neurons. 
We experimentally evaluate the improved architecture in various settings: Minimal arithmetic tasks, one-layer calculations where among others the relevant inputs have to be recognized and simple function learning where a combination between operations has to be learned in two layers. 

Our main contributions are the improvement of the extrapolation results of the NALU and the mixed-signed multiplication with negative values as result.

\textbf{Structure.}
The paper is structured as follows:
The next section describes related work. 
Section~\ref{sec:inalu} explains the NALU and our improved model iNALU in more detail. 
Experiments are presented in Section~\ref{sec:experiments} and the results are discussed in Section~\ref{sec:discussion}.
Finally, Section~\ref{sec:conclusion} concludes the paper.

%% file: chapters/related.tex
\section{Related Work}
\label{sec:relatedwork}

This section reviews related work on processing mathematical operations using neural networks. 

Kaiser and Sutskever~\cite{kaiser2015neural} present Neural GPU, a neural network architecture which is able to solve algorithmic tasks. 
The architecture of Neural GPU is based on a type of convolutional gated recurrent units (CGRU). 
The authors show that their approach is able to learn long binary summations and multiplications and that their approach generalizes well for longer numbers. 
However, in the experimental evaluation, the input to the network is limited to four symbols. 
Freivalds and Liepins~\cite{freivalds2017improving} propose an improvement for the Neural GPU which speeds up the training time and provides better generalization.  
Similarly, Kalchbrenner et al.~\cite{kalchbrenner2015grid} propose Grid Long Short-Term Memory, a network of LSTM cells which is able to add 15-digit integer numbers. 
These three approaches from Kaiser and Sutskever~\cite{kaiser2015neural}, Freivalds and Liepins~\cite{freivalds2017improving} and Kalchbrenner et al.~\cite{kalchbrenner2015grid} process sequential data and are able to learn simple algorithmic tasks. 

Another work in this area is proposed by Chen et al.~\cite{chen2018neural}. 
The authors use reinforcement learning to solve mathematical operations such as summation, subtraction, multiplication or division. 
However, compared to our setting, Chen et al. provide the mathematical operation as an additional input to their network.

The most similar work to ours is from Trask et al.~\cite{trask2018neural}. 
The authors propose the neural arithmetic logic unit which is able to perform mathematical operations. 
They show in their experimental evaluation that their model generalizes better than traditional neurons for extrapolation tasks. 
However, the NALU has some limitations which we will discuss in Section \ref{sec:chall}. 

Other works with small intersections are by Zaremba and Sutskever  \cite{zaremba2014learning} as well as by Reed and de Freitas~\cite{reed2015neural}. 
Both use Recurrent Neural Networks to execute small code snippets which contain the summation of digits.
Counting the number of specific objects in images can also be seen in the wider scope of related work. 
In this context, works by Xie et al.~\cite{xie2018microscopy} and Zhang et al.~\cite{zhang2015cross} involve counting the number of microscopy cells respectively crowd counting. 

%% file: chapters/nalu.tex
\section{Improved Neural Arithmetic Logic Unit}
In this chapter, we first describe the Neural Arithmetic Logic Unit and discuss properties and challenges. 
We then introduce iNALU, a new model variant, to address these challenges.

\label{sec:inalu}

\begin{figure*}[bth]
	\includegraphics[width=1\textwidth]{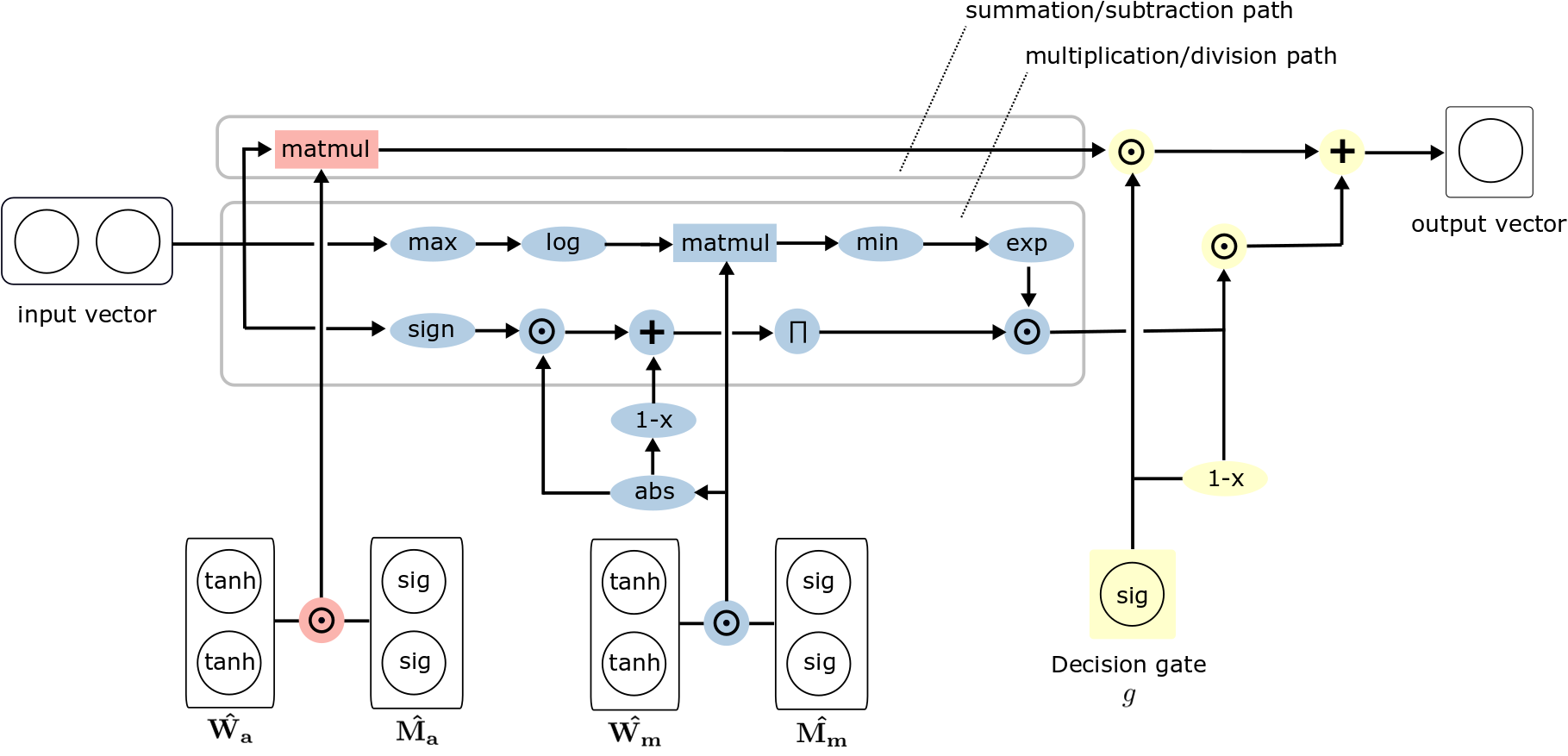}
	\centering
	\caption{Architecture of the improved Neural Arithmetic Logic Unit (iNALU).}
	\label{fig:inaluCom}
\end{figure*}

\subsection{Neural Arithmetic Logic Unit}
The NALU as proposed in \cite{trask2018neural} consists of a multiplicative and a summative path, which can be seen as a linear layer with a weight matrix constrained to $[-1, 1]$. The weights $\mathbf{W}$ are constructed as point-wise product between a matrix $\mathbf{\hat{W}}$ with $\tanh$ activations and a matrix $\mathbf{\hat{M}}$ with $\text{sigmoid}$ ($\sigma$) activations.

\begin{equation}
\label{eq:weigthmatrix}
\mathbf{W} = \tanh(\mathbf{\hat{W}}) \odot \sigma(\mathbf{\hat{M}})
\end{equation}

By matrix-multiplication of inputs $\mathbf{x}$ and weights $\mathbf{W}$, output values stay within the magnitude of the input values (since $-1 \leq \mathbf{W}_{i,j} \leq 1$) and result in the summation for values of $\mathbf{W}_{i,j} = 1$ and subtraction for values of $ \mathbf{W}_{i,j} = -1$. By balancing the weights between $-1$, $0$, and $1$ any function composed of adding, subtracting and ignoring inputs can be learned. This summative path $\mathbf{a}$ is defined in Equation~\ref{eq:summation}.

\begin{equation}
\label{eq:summation}
\mathbf{a} = \mathbf{xW}
\end{equation}

To multiply or divide, this calculation is performed in $\log$-space (see Equation~\ref{eq:multiplication}).
The NALU encounters the problem of calculating $\log(x)$ for $x \leq 0$ by restricting the calculation to absolute input values and adding a small constant value $\epsilon$.

\begin{equation}
\label{eq:multiplication}
\mathbf{m} = \exp \left(\log(|\mathbf{x}| + \epsilon)\mathbf{W} \right)
\end{equation}

A gate is used to decide between the summative and the multiplicative path depending on the input vector.

\begin{equation}
\label{eq:gate}
g = \sigma(\mathbf{xG})
\end{equation}

Since the gate-weights $\mathbf{G}$ are multiplied with the inputs $\mathbf{x}$, each gate dimension maps to an input dimension and contains the corresponding weight to which the input shall contribute to the decision between both arithmetic paths.

The output is obtained by adding the gated summative (see Equation~\ref{eq:summation}) and multiplicative (see Equation~\ref{eq:multiplication}) paths.
\begin{equation}
\label{eq:origNalu}
\mathtt{NALU}_o\mbox{: }  \mathbf{y_{\text{nalu}}} = g \cdot \mathbf{a} + (1 - g) \cdot \mathbf{m}
\end{equation}

The NALU model can finally be implemented in two ways.
One can either use a weight vector $\mathbf{G}$ and a scalar gate $g$ or a weight matrix $\mathbf{G}$ and a gate vector $\mathbf{g}$. Tasks for which the selection of the operation is different for each output or for which it is depending on input values  might benefit from the gate matrix. However, this introduces additional parameters which for many tasks are unnecessary. %In conjunction with mixed sign gating this can lead to non-convergence of the model. 
In our experiments we use both, vector based NALU and a NALU with matrix based gating for comparison. \label{sec:matrixvectornalu}

However, some of these design decisions for the NALU result in challenges we want to address in the following section.

\subsection{Challenges}
\label{sec:chall}

\subsubsection{Exploding Intermediate Results}
In our experiments, we observe that training often fails because of exploding intermediate results especially when stacking NALUs to deeper networks and having many input and output variables.
For example, consider a model consisting of four NALU layers with four inputs, four outputs neurons each and a simple summation task. Assuming the same magnitude for all input dimensions the first layer could (depending on the initialization) calculate $x^4$ for each output dimension whereas the following layer could calculate $(x^4)^4$ ultimately leading to $x^{4^l}$ for layer $l$.
Therefore, the calculation can exceed the valid numeric range already in the forward pass ultimately causing the training to fail. For example in a network with three NALU layers in a MNIST classification downstream task, the NALU models failed after the first training steps (resulting in NaNs).

\subsubsection{Multiplication / Division with Negative Result}

The NALU by design isn't capable of multiplying or dividing values with a negative result.
In the multiplicative path, the input values are represented by their absolute value to guarantee a real-valued calculation in log-space. Therefore, learning multiplication for mixed signed data with a result $y < 0$ fails. Since the NALU is expected to learn either multiplication\,/\,division or summation\,/\,subtraction in each layer, $\text{sign}(y)$ is in the multiplicative case clearly determined by the number of negative multiplicands being even or odd. 
Since input dimensions can be deactivated for $ \mathbf{W}_{i,j} = 0$, the sign can't be inferred counting negative input variables. In the next section, we propose a method taking deactivated input dimensions into account to correct the sign of the multiplicative path.

\subsubsection{Mixed Sign Gating}

Despite the summative path is capable of dealing with mixed input signs, the construction of the gating mechanism leads to problems.
If input values are constantly positive or constantly negative, Equation~\ref{eq:gate} leads to the desired gating behavior. However, if the input values mix negative and positive values, $\sigma$ and thus the gate is dependent of the sign since $\mathbf{G}$ can't fit the designated gate state systematically correctly. 

\subsubsection{Initialization Sensitivity}

We observed that the NALU architecture is very prone to non-optimal initializations, which can lead to vanishing gradients or optimization into undesired local optima. Finding the optimal initialization in general is difficult since it depends on the task and the input distribution, which in a real world scenario is both unknown.

\subsubsection{Leaky Gates}

Another challenge we observe are variables, not tied near to their boundaries.
Generally in the NALU design the variables $\mathbf{W}$ and $g$ are intended to reach their boundaries of $[-1, 1]$ and $[0, 1]$ for maximum precision. However, during training and for interpolation, an approximation of the intended calculation having gates trained to $g = 0.5$, for example with a specific configuration of $\mathbf{W}$ represents a local optimum. For extrapolation such a model fails by large margin. We suggest regularizing the trained variables to avoid this behavior.

\subsection{Improvements}
\label{sec:extensions}
This section describes the improvements we incorporate in our iNALU model to address the aforementioned challenges. Figure \ref{fig:inaluCom} summarizes the complete model architecture. In the following, we discuss each improvement and extension in detail.

\subsubsection{Independent Weights}  

The summative and the multiplicative paths share their weights $\mathbf{\hat{W}}$ and $\mathbf{\hat{M}}$ in the NALU model. We propose using separate weights for each path for two reasons: First, the model can optimize $\mathcal{\mathbf{W}}$ for the multiplicative and summative path without interfering the other path. 
For example, in a setting with inputs $a, b < -1$ with the operation $a \times b$, the result would be a positive number greater than $1$ and the optimal parameter setting would be $\mathbf{W}_a = \mathbf{W}_b = 1$ and $g=0$. 
However, the only way for the summative path (see Equation~\ref{eq:summation}) to generate positive results is to force the weights $\mathbf{W}_a$ and $\mathbf{W}_b$ towards $-1$.
In this case, the summative and multiplication path force the weights into opposite directions.
With separate weights, the model can learn optimal weights for both paths and select the correct path using the gate. 
Second, consider the multiplicative path yields huge results whereas the summative path represents the correct solution but yields relatively small results. 
In that case, the multiplicative path influences the results even if the sigmoid gate is almost closed. 
For example in a setting with inputs $a,b,c > 0$ with the desired result $a+b$, the summative path yields the correct solution and the optimal weight setting is $\mathbf{W}_a = \mathbf{W}_b = 1$, $\mathbf{W}_c = 0$ and $g=1$. 
In that case, $\mathbf{W}$ may contain very small weights to omit the input $c$. 
However, small negative weights for $\mathbf{W}_c$ (e.g. $-1e-5$) will lead to the situation, that the multiplication path divides the inputs $a$ and $b$ by values near to $0$ which results in large numbers. 
Consequently, the multiplicative path influences the results even if the gate (see Equation~\ref{eq:gate}) is almost closed.  
In this case, the model with independent weights can optimize $\mathbf{W_{\text{m}}}$ to smaller values to mitigate influence caused by the leaky gate. Our modifications are summarized in the following equations:

\begin{align}
\label{eq:wa}
\mathbf{W_{\text{a}}} &= \tanh(\mathbf{\hat{W}}_a) \odot \sigma(\mathbf{\hat{M}}_a) \\
\mathbf{W_{\text{m}}} &= \tanh(\mathbf{\hat{W}}_m) \odot \sigma(\mathbf{\hat{M}}_m) \\
\mathbf{a} &= \mathbf{xW_{\text{a}}} \\
\mathbf{m} &= \exp \left( \log(|\mathbf{x}| + \epsilon)\mathbf{W_{\text{m}}} \right) \label{eq:wm}
\end{align}

\subsubsection{Weight and Gradient Clipping} To address the challenge of exploding intermediate results in a multi-layer setting, we improve the  model by clipping exploding weights in the back-trans\-formation from $\log$-space (see Equation~\ref{eq:clipping}), which improves the stability of deep networks. To validate this, we incorporated three NALU layers in a MNIST classification downstream task. Our proposed clipping mechanism resulted in successful training solving the task very well\footnote{with an accuracy of $0.94$ after 64000 steps}, whereas the original NALU fails.

\begin{equation}
\label{eq:clipping}
\mathbf{m} = \exp \Big( \min\big(\log\left(\max(|\mathbf{x}|, \epsilon)\right)\mathbf{W}, \omega \big) \Big)
\end{equation}

Further, we apply gradient clipping to avoid stability problems due to large gradients, which can for example occur when input values are near to zero. We set $\epsilon$  to $10^{-7}$ and $\omega$ to 20.

\subsubsection{Sign Correction} The NALU cell by design isn't capable of multiplying or dividing values with a negative result. Therefore, NALU fails calculating multiplication of mixed signed data.
We propose a solution by taking the sign of relevant input values into account (i.e., all $\mathbf{W}_{i,j} \neq 0$).

\begin{align}
\mathbf{msm} &= \sign(\mathbf{x}) \odot |\mathbf{W}| + 1 - |\mathbf{W}| \label{eq:msm}\\ 
\mathbf{msv} &= \prod\limits_j \mathbf{msm}_{ij} \label{eq:msv} \\
\mathtt{NALU}_s\mbox{: }  \mathbf{y_{\text{nalu}}} &= g \cdot \mathbf{a} + (1 - g) \cdot \mathbf{m}\cdot\mathbf{msv}
\end{align}

The multiplication sign matrix $\mathbf{msm}$ (see Equation~\ref{eq:msm}) contains values in the range $[-1, 1]$. If $\mathbf{W}$ is discrete i.e. $\mathbf{W}_{i,j} \in \{-1, 0, 1\}$, which is a desired property \cite{trask2018neural} to achieve generalization and interpretability, $\mathbf{msm}$ is also discrete, i.e. $\mathbf{msm}_{i,j} \in \{-1, 1\}$.
By multiplying the columns of $\textbf{msm}$, we get the sign vector containing the correct sign for the multiplication path (see Equation~\ref{eq:msv}).

\subsubsection{Regularization} In general, $\mathbf{W}$ and $g$ having discrete values is often crucial for a model to generalize and learn a calculation correctly instead of approximating the solution. This becomes even more important for the sign corrected multiplication. We therefore propose regularizing the weights such that $\mathbf{\hat{W}}$, $\mathbf{\hat{M}}$ and $\mathbf{G}$ don't contain values near zero by introducing a piecewise linear regularization term (see Equation~\ref{eq:reg}) which adds to the loss until the weight has reached a discretization threshold $t$. We found $t = 20$ suitable since $\sigma(-20) < 10^{-9}$.

\begin{equation}
\label{eq:reg}
\mathcal{L}_{\text{reg}}(w) = \frac{1}{t} \max(\min(-w, w) + t, 0)
\end{equation}

 Note that the regularization can cause gradient-directions contradicting the gradient-direction of the loss without regularization depending on the initialization. We try to mitigate this problem by incorporating the regularization only after several training steps, when the loss is below a threshold (see Section \ref{sec:experiments} for more details).
 
 Further, regularization is especially useful to improve extrapolation performance. For example, we evaluate regularization in the Simple Function Learning Task (see Section \ref{sec:simplefunc}) setup for a summation task (i.e. an overdetermined task where an optimal and generalizing solution can be found even for  $-1 < \mathbf{W}_{i,j} < 1$). We obtained after 10 epochs without regularization an interpolation loss of $5.95\cdot 10^{-4}$ and an extrapolation loss of $4.46\cdot 10^{11}$. The model has found a suitable approximation for the training range but failed to generalize. Introducing regularization after the 10th epoch, after 5 more training epochs we reach an interpolation loss of $2.2\cdot 10^{-13}$ and an extrapolation loss of $2.2\cdot 10^{-11}$, whereas without regularization we just improve the interpolation loss ($8.30\cdot 10^{-5}$) and the extrapolation loss even impairs ($8.76\cdot 10^{14}$).

\subsubsection{Reinitialization} Since NALU doesn't recover well from local optima by its own \cite{madsen2019measuring}, we suggest a reinitialization strategy. 
This strategy evaluates the loss for each $m$-th epoch and randomly reinitializes all weights if the loss did not improve for the last $n$ steps and if the loss is greater than a predefined threshold.

\subsubsection{Independent Gating}
In many tasks, the decision which operation path to choose is not depending on the input values but instead fixed for the task, e.g., typical spreadsheet tasks like calculating the sum or product of different columns.
For this case we propose a model, where the scalar gate is replaced by a vector. In our model this vector is independent from the input neurons (see Equation~\ref{eq:indgate}) and only trained to fit the gates to the task. 

\begin{equation}
\label{eq:indgate}
\mathbf{g_i} = \sigma(\mathbf{G})
\end{equation}

Note that choosing a vector over a scalar enables our model to select the operation for each output independently.

%% file: chapters/experiments.tex
\section{Experiments}

\label{sec:experiments}

\subsection{Prerequisites}
This section describes at first the general commonalities of all experiments.

\textbf{Datasets.} For all experiments, we evaluate on an interpolation task as well as an extrapolation task. 
For the interpolation task, the training and evaluation dataset are drawn from the same distribution.
For the extrapolation task, the evaluation dataset is drawn from a distribution with a different value range in order to evaluate the  ability to generalize.
Each dataset contains $N=64\,000$ samples.

\textbf{Tasks.} For our experiments we focus on mathematic operations since these are the building-blocks of more complex tasks. All tasks involve applying a operation $\diamond \in \{+,-,\times, \div\}$ to input and/or hidden variables $a$ and $b$ to calculate $y = a~\diamond~b$. Note that \cite{trask2018neural} introduces additional operations such as identity, square and the square-root but since these operation are special cases of the basic operations, their learning performance is closely correlated with the performance on the basic operations and therefore omitted for the sake of clarity. The input variables for all experiments are sampled randomly from a distribution $\mathcal{P}$ with a parameterization $\lambda$, which are defined in the following sections in more detail. 
Note that for $\mathcal{P}=\mathcal{N}$ the normal distribution for our experiments is truncated to $\lambda = [a,b] = [\mu - 3\sigma, \mu + 3 \sigma]$ (containing $\approx 99,7\%$ probability mass) to ensure that the extrapolation task is performed out of the test distribution range. For the exponential distribution ($\mathcal{P}=\mathcal{E}$) the extrapolation task involves no extrapolation in a literal sense but rather examines if generalization for different $\lambda$ values can be achieved.

\textbf{Evaluation.} In contrast to \cite{trask2018neural}, we choose a different evaluation strategy: Trask et al. reported the error for each operation relatively in comparison to a random initialized network prior training. Since the performance of the untrained network is constantly bad, the relative performance reported this way can be used to decide how well each architecture performs rank-wise but it can't be used to infer, to which extend the calculated result differs from the expected result. 
Instead, we use a more intuitive approach for evaluation and report the mean squared error (MSE) between the calculated and the expected results over the complete evaluation datasets. For all experiments we report results for extrapolation, since this is the more difficult task.

\begin{equation}
\text{MSE}(y^\text{pred}, y^\text{real}) := \frac{1}{N}\sum^N_i (y_i^\text{pred} - y^\text{real}_i)^2
\end{equation}

The MSE comes along with another advantage. Combined with a predefined threshold, the MSE can be used to evaluate if the model reaches the necessary precision~\cite{madsen2019measuring}. If not stated otherwise we understand a MSE $ \leq 10^{-4}$ as successful training.

We repeat each experiment ten times with different random seeds. This procedure examines if the performance is stable or how much it scatters randomly.

\textbf{Training.} We use the Adam optimizer \cite{kingma2014adam} in mini-batch training with a learning-rate of $0.001$ and a batch size of $64$.
Training is done for $100$ epochs using the MSE as loss.
Clipping, regularization and random reinitialization as described in Section \ref{sec:extensions} are implemented.
Regularization is activated after 10 epochs whenever the training loss  $\mathcal{L} < 1$. 
Reinitialization is applied each 10th epoch if the loss hasn't improved over $m=10000$ steps. 
This means during training reinitialization can occur up to nine times. Note that this method could lead to incompletely trained models if a reinitialization occurs late during training in favor of a fair model comparison.

\subsection{Experiment 1 - Minimal Arithmetic Task}
\label{sec:minimal}

 Experiment 1 constructs the most minimalistic task where the model has two inputs and one output and analyzes the influence of the input value distribution by sampling $a$ and $b$ from uniform, truncated normal and exponentially distributed random variables in various ranges. 

\begin{figure}[tb!]
\centering

\includegraphics[width=0.95\textwidth]{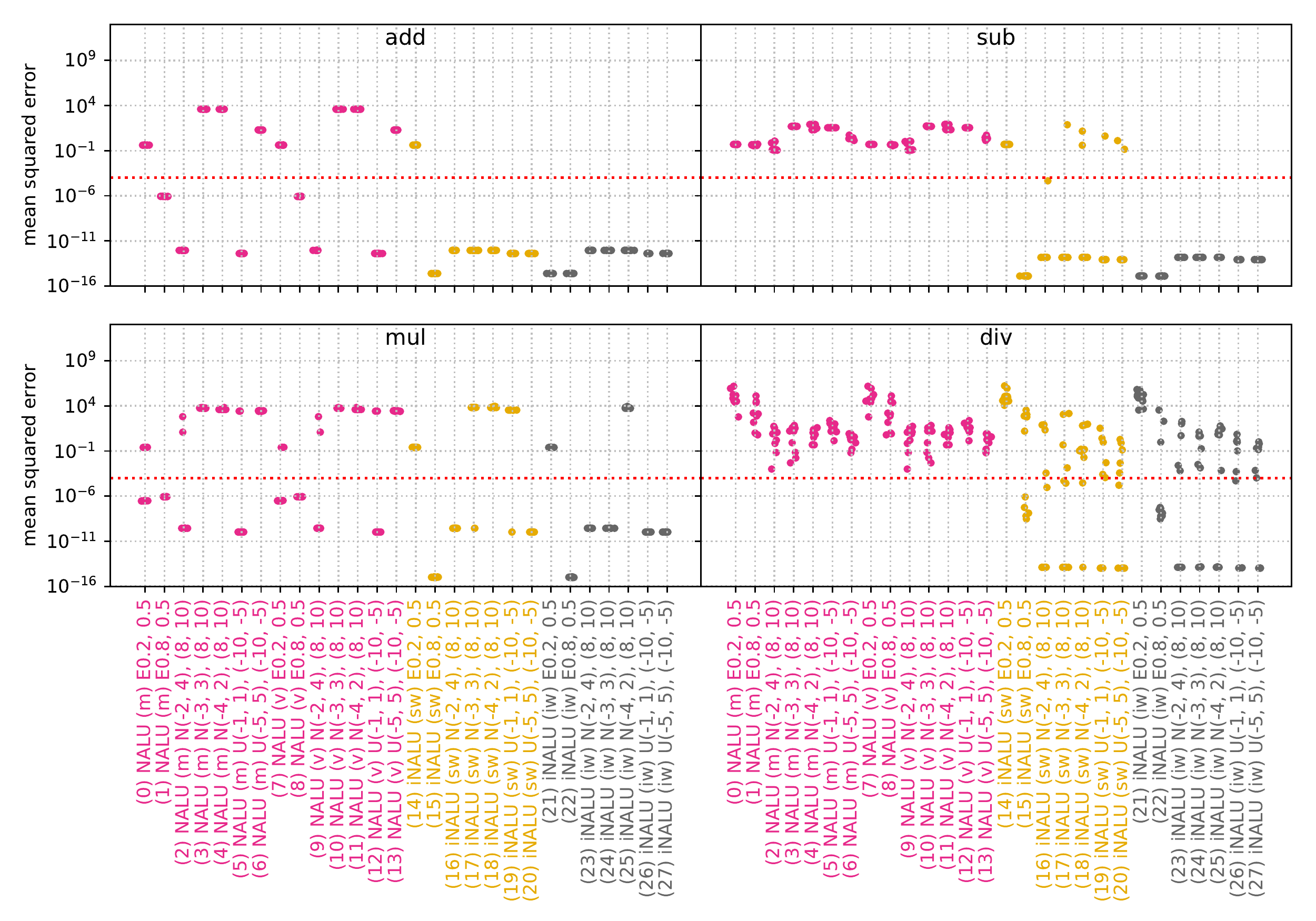} 

	  \caption{\label{fig:ex1ex}MSE for various input distributions per operation over the extrapolation test dataset of experiment 1 (minimal arithmetic task). The original NALU is colored in pink, (m) stands for the matrix gating, and (v) for the vector gating version. Our iNALU models are depicted in yellow for the shared weights variant and grey for the version with independent weight matrices for the summative and multiplicative path. For truncated normal (N) as for uniform distributed data (U), the first parameter tuple represents the training data range, the second tuple represents the extrapolation range. For exponentially distributed data (E) the parameter $\lambda$ is reported.} 
\end{figure}

\begin{figure}[bht!]

\includegraphics[width=0.95\textwidth]{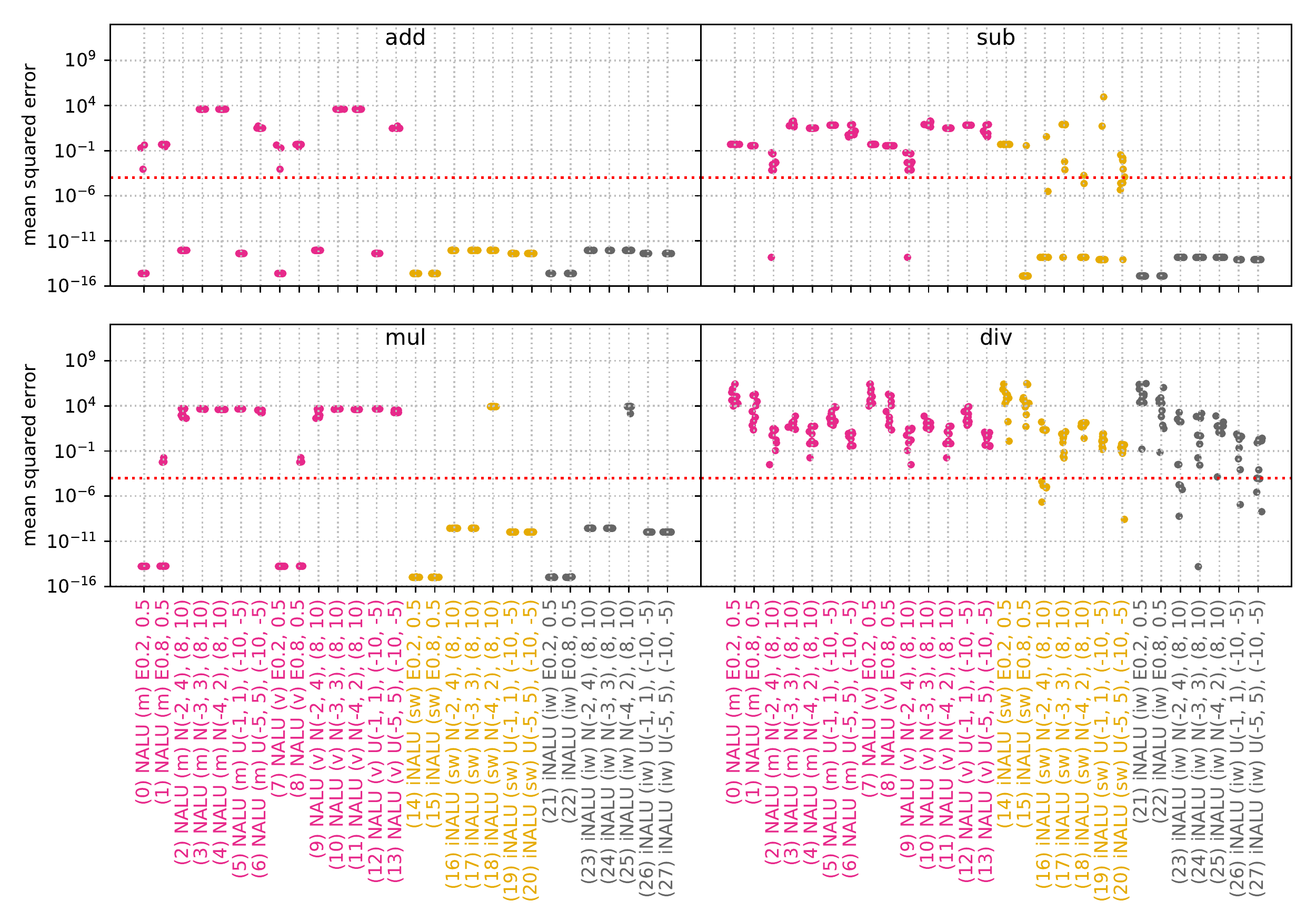}

	  \caption{\label{fig:ex2ex}MSE for various input distributions per operation over the extrapolation test dataset of experiment 2 (simple arithmetic task). For a detailed description see Fig.~\ref{fig:ex1ex}.
	  }
\end{figure}

\textbf{Results.} The extrapolation results of this experiments are presented in Figure \ref{fig:ex1ex}.

In general our iNALU models perform substantially better on all operations. With the exception of exponentially distributed data for $\lambda = 0.2$, for summation all and for subtraction almost all models succeed. For multiplication iNALU with independent weights performs best reaching very good precision with the exception of $E(0.2)$ and $N(-4,2)$. All models yield worse results for division.
In fact, for the original NALU, no tested input parameter configuration leads to acceptable MSEs (the average MSE is $4.36 \cdot 10^{4}$). 
Our models also yield mixed results, some solving the task nearly perfect after one to six reinitialization but others failing after nine reinitialization as well.

\subsection{Experiment 2 - Simple Arithmetic Task}
Experiment 2 is a generalization of the minimal arithmetic task where the model has to learn to ignore irrelevant input dimensions to calculate the correct solution.

This setting is motivated by real world tasks like spreadsheet calculations where one column is calculated by applying a simple operation to two specific columns while other columns are present but must not influence the result.

The model consists of one NALU layer with ten inputs and one output. We test the same input distributions as in the minimal arithmetic task (see \ref{sec:minimal}). 

\textbf{Results.} 
Figure \ref{fig:ex2ex} shows the results of this experiment.	Although, the setting of experiment 2 is slightly more complex than experiment 1, most performance patterns repeat. In the following, we want to highlight some interesting exceptions. 

For input data sampled from an exponential distribution, the results improve for the original NALU models especially for summation and multiplication. 
For summation training is unstable, since some models succeed but others fail to learn the task. 
In contrast to the minimal arithmetic task, iNALU succeed for summation of exponentially distributed data with $\lambda = 0.2$ and shows better results for multiplication. 
For division the situation of unstable training as discussed before even worsens such that only very few of our iNALU models succeed ($\approx 6.4\%$ of all experiments reach a MSE $< 10^{-5}$). 
The original NALU failed constantly for division. 
For subtraction, our model with shared weights is slightly more unstable but our model with independent weights still yields stable results and calculates precisely.

\subsection{Experiment 3 - Influence of Initialization}
Experiment 1 suggests that training is unstable for some operations (subtraction and division).
Whereas some of our improved models happen to solve the minimal task flawlessly, others fail to converge. 
As a consequence, suitable initialization seems to be crucial for successful training of more complex architectures. This fact is also confirmed by Madsen and Johansen~\cite{madsen2019measuring}.

In this experiment, we analyze the effect of different parameters for random weight initialization of the neurons.

In contrast to the Minimal Arithmetic Task, the variables $a$ and $b$ are constructed by summing up 100 input vector entries assigned to $a$ and $b$. Since \cite{trask2018neural} doesn't specify the assignment in detail, we construct it by randomly assigning entries mutually exclusive to $a$ and $b$ and demand some inputs to be ignored by the model (since they neither contribute to $a$ nor to $b$). We decide on the assignment once per task randomly such that the assignment is constant for all samples. Note, that the assignment is not an additional input to the neural network but instead it has to learn this assignment.

For this study, we examine the model performance of our iNALU model with shared weights for standard normal distributed input values such that $\mathcal{P} = \mathcal{N}$ and $\lambda = (\mu, \sigma) = (0, 1)$. We choose to initialize the model weights following a normal distribution as well. To find suitable initialization parameters, we performed an exhaustive search for the parameters $\mu_g, \mu_{\hat{M}}, \mu_{\hat{W}} \in \{-1, 0, 1\}$ and $\sigma_g, \sigma_{\hat{M}}, \sigma_{\hat{W}} \in \{0.1, 0.5\}$. %An initialization of $-1$ for $g_{i,j}$ for example yields $\sigma(-1) \approx 0.269$. 
We repeat each parameter setting 20 times with different seeds to be able to asses the model stability.
Note that too large initializations bias the model towards specific operations, but especially sigmoid activations suffer from small random initializations \cite{glorot2010understanding}.

\textbf{Results.} Table \ref{tab:sfinit} shows the results of our parameter search. We consolidated the results for $\sigma = 0.1$ and $\sigma = 0.5$, since both parameters yielded similar results and report the maximum MSE of all runs for each parameter setting. This is a very strict evaluation metric since only 1 of 20 models failing could obfuscate 19 successful runs. However, we are particularly interested in parameters which lead to stable models. The results support our finding from the arithmetic experiments that division is very unstable to learn. To be precise, no model solved the problem for all parameter configurations and repetitions.
Stable parameter configurations could be found for the remaining operations.
Overall the configuration $(\mu_g, \mu_{\hat{M}}, \mu_{\hat{W}}) = (0, -1, 1)$ is clearly most stable among all tested parameters for this task and architecture.

\subsection{Experiment 4 - Simple Function Learning Task}
\label{sec:simplefunc}
For the Simple Function Learning Task, we keep the setting of the previous experiment but focus on the comparison of our model using both, combined path-weights and separated path-weights to the originally proposed NALU in both variants (see section \ref{sec:matrixvectornalu}).

Since we found suitable initializations, we sample from uniform and truncated normal distribution and interpolate within the interval $[a,b] = [-3, 3]$ for both. This translates to a standard normal distribution ($\mu = 0, \sigma = 1$) for the truncated normal distribution. For the extrapolation interval we choose $[3,4]$ and $[-5, -3]$ to test positive as well as negative values outside the training range with different standard deviations.

\begin{figure}[tb!]
\includegraphics[width=0.95\textwidth]{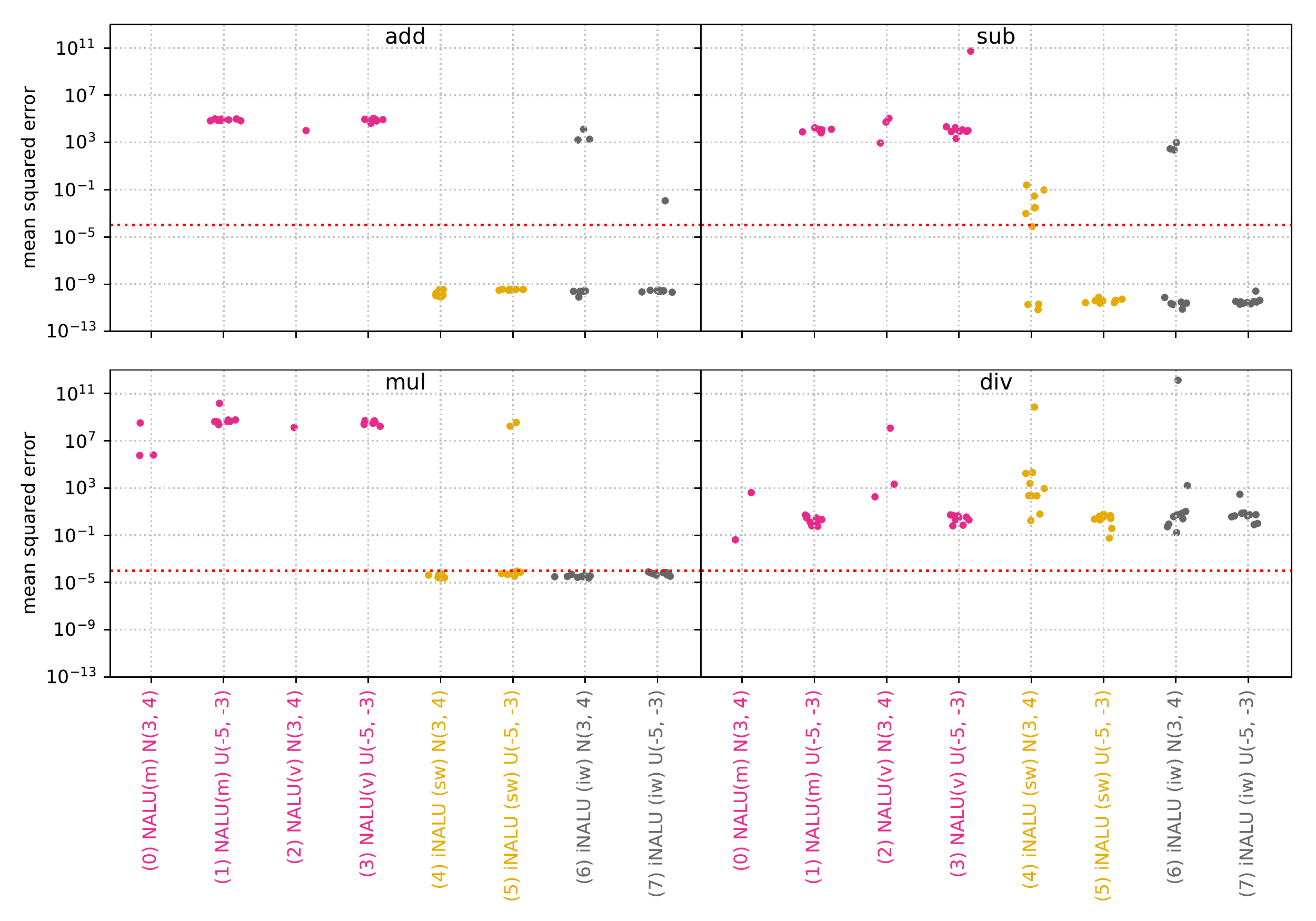}
	\centering
	 %\vspace{-0.7cm}
	  \caption{Extrapolation MSE for Experiment 4 (Simple Function Learning Task). Original NALU with gating matrix (m) and gating vector (v) are colored pink, our iNALU model with shared weights (sw) is colored yellow and with independent weights (iw) in grey.}
  \label{fig:ex4}
 
\end{figure}

\textbf{Results.} Figure \ref{fig:ex4} shows, that our iNALU models outperforms the original NALU for summation, subtraction and multiplication on almost all runs. Our model with independent weights is most promising since almost all runs succeed. However, few outliers indicate that the stability problem is not completely solved yet. This especially holds for division where all models fail to learn the operation correctly.

\sisetup{round-precision=2} 

\newcommand{\csvautotabularcenter}[2][]{\csvloop{autotabularcenter={#2},#1}}

\pgfplotstableread[col sep=comma]{param-simple-func.csv}\mytable %Read the table
\pgfplotstableadduniquecol{eg}{\mytable} % Add the unique column
\pgfplotstableadduniquecol{em}{\mytable} % Add the unique column
\pgfplotstableadduniquecol{ew}{\mytable} % Add the unique column

\pgfplotstableset{
    col style/.style 2 args={
        postproc cell content/.append code={%
            \count0=\pgfplotstablecol
            \advance\count0 by1\relax
            \ifnum\count0=#1
                \pgfkeysalso{#2}%
            \fi
        }
    }
}

\begin{table}
\caption{Maximum MSE over all models for the Simple Function Learning Task (extrapolation) for weight initializations means of $-1$, $0$, $1$. Successful configurations (maximum loss $< 0.001$) in bold, percentage of successful repetitions in brackets.}
\vspace{1em}
\label{tab:sfinit}
\scalebox{0.94} {
\pgfplotstabletypeset[string type,
    every head row/.style={before row=\toprule, after row=\midrule},
    every last row/.style={after row=\bottomrule},
    every row no 8/.style={after row=\midrule},
    every row no 17/.style={after row=\midrule},
    every row no 2/.style={after row={\cmidrule{2-7}}},
    every row no 5/.style={after row={\cmidrule{2-7}}},
    every row no 11/.style={after row={\cmidrule{2-7}}},
    every row no 14/.style={after row={\cmidrule{2-7}}},
    every row no 20/.style={after row={\cmidrule{2-7}}},
    every row no 23/.style={after row={\cmidrule{2-7}}},
    columns={u-eg,u-em,u-ew,ADD,DIV,MUL,SUB},
    columns/u-eg/.style={column name=$E[\mathbf{G}]$},
    columns/u-em/.style={column name=$E[\mathbf{\hat{M}}]$},
	columns/u-ew/.style={column name=$E[\mathbf{\hat{W}}]$},
    col style={4}{@cell content=#1},
    col style={5}{@cell content=#1},
    col style={6}{@cell content=#1},
    col style={7}{@cell content=#1},
    every first column/.style={ column type/.add={@{}}{} }, % this does not work
every last column/.style={ column type/.add={}{@{}} }, % this does not work
]{\mytable}
}
\end{table}

\section{Discussion}

\label{sec:discussion}

The experiments in Section~\ref{sec:experiments} analyzed the ability of the original NALU and our iNALU to solve various mathematical tasks and show that the performance of the NALU heavily depends on the distribution of the input data. 
The quality of the iNALU also depends on the input distribution but is in general more stable and achieves better results. 
Experiment 2 extends the arithmetic task by switching off several inputs. 
The results reinforce the findings of the first experiment that iNALU achieves better and more stable results than NALU. 
The differences between both iNALU models can be explained by the separate weighting matrix for summation/subtraction and multiplication/division. 
In Experiment 4, the iNALU achieves for three of four operations acceptable results whereas the original NALU fails for all four operations. 

In general, the MSE calculated on the extrapolation datasets provides a good intuition if the NALU has learned the correct logical structure which is resilient to other value ranges. 
The interpolation results are very similar regarding the relative performance of all models but in general achieve a higher precision and thus a lower MSE (e.g. for summation in experiment 1 our iNALU model with independent yields $6.14\cdot10^{-15}$ for interpolation and $5.45 \cdot 10^{-13}$ for extrapolation on average MSE).

Further, experiments 1, 2 and 4 show that the operation division is the most challenging task for NALU and iNALU. 
The instabilities for division might be explained by the special case of dividing by near-zero and the sampling strategy for $a$ and $b$: For sampling inputs in an interval including 0, division might cause huge or very small results depending on the assignments of dividend or divisor which are represented by completely different weights. Possibly irrelevant input variables might therefore influence the result by such magnitude that there is no clear gradient signal for the assignment.

Another observation is that the optimal initialization is dependent on many factors such as task, model size and value range. 
We want to emphasize that our parameter study is not intended to raise a claim for generally finding the optimal parameters, but rather to find initialization parameters for this specific task to allow a model comparison. 
Our study suggests the parameter configuration $(\mu_g, \mu_{\hat{M}}, \mu_{\hat{W}}) = (0, -1, 1)$ which seems to be reasonable, since it treats the summative/subtraction path and multiplicative/division path equally at beginning and assigns small activation weights to all inputs. 
We believe that the problem of generally finding optimal or near optimal initializations is an interesting and theoretically challenging task for future work.

%% file: chapters/conclusion.tex
\section{Conclusion}
\label{sec:conclusion}

Recently, the NALU architecture was proposed to learn mathematical relationships, which are necessary for solving various machine learning tasks. In this paper, we proposed an improved version of this architecture called iNALU.
The original NALU is only able to calculate non-negative results for multiplication and division by design and often fails to converge to the desired weights. 
We solved the issue of multiplying and dividing with mixed-signed results and proposed architectural variants for shared and independent weights with input independent gating.
Further, we introduced a regularization term and a new reinitialization strategy which help to overcome the problem of unstable training.

We evaluated the improvements on four large scale experiments which examine the influence of different input distributions and task-unrelated inputs. 
The first two experiments analyze the basic capabilities of NALU and iNALU. 
Further, the parameter study for the Simple Function Learning Task shows that the choice of weight initializations has a huge impact on model stability.
The parameter study revealed suitable initialization parameters. We showed that our proposed architectures can learn simple mathematical functions and outperforms the reference models in terms of precision and stability.

Future work encompasses analyzing the stability issue from a theoretical point of view and evaluating the extensions in various downstream tasks.
Last but not least, we want to improve the division in more complex learning scenarios.